\documentclass[journal]{IEEEtran}
\usepackage{amsfonts}
\usepackage{cite,graphicx,amsmath,amsthm}
\usepackage{subfigure}
\usepackage{fancyhdr}
\usepackage{dsfont}
\usepackage{array,color}
\usepackage{bm}
\usepackage{float}
\usepackage{algorithm}
\usepackage{algpseudocode}
\usepackage{multirow}
\usepackage{bbm}
\usepackage{graphicx}
\usepackage{float}
\usepackage{subfigure}
\usepackage{ragged2e}
\usepackage{booktabs}
\usepackage{bbding}
\usepackage{multirow}
\usepackage{makecell}
\usepackage{amsmath,amssymb,amsfonts}
\usepackage[]{caption2}
\usepackage[table,xcdraw]{xcolor}
\usepackage{cite}

\newtheorem{theorem}{Theorem}

\newtheorem{lemma}{Lemma}

\newtheorem{proposition}{Proposition}

\newtheorem{corollary}{Corollary}

\newtheorem{property}{Property}

\newtheorem{remark}{Remark}

\newtheorem{claim}{Claim}

\allowdisplaybreaks[4]

\begin{document}
\title{{Distributed Dynamic Map Fusion via Federated Learning for Intelligent Networked Vehicles}}

\author{Zijian Zhang$^{1,5,\dag}$, Shuai Wang$^{1,2,3,\dag}$, Yuncong Hong$^{2}$, Liangkai Zhou$^{2}$, and Qi Hao$^{1,3,4,*}$

\thanks{
This work has been presented in 2021 IEEE International Conference on Robotics and Automation (ICRA).
This work was supported in part by the Science and Technology Innovation Committee of Shenzhen City under Grant JCYJ20200109141622964,
in part by the Intel ICRI-IACV Research Fund (CG\#52514373), and in part by the National Natural Science Foundation of China under Grant 62001203.

$^{\dag}$Equal contribution.

$^{*}$Corresponding author: Qi Hao (hao.q@sustech.edu.cn).

$^{1}$Department of Computer Science and Engineering, Southern University of Science and Technology, Shenzhen 518055, China.

$^{2}$Department of Electrical and Electronic Engineering, Southern University of Science and Technology, Shenzhen 518055, China.

$^{3}$Sifakis Research Institute of Trustworthy Autonomous Systems, Southern University of Science and Technology, Shenzhen 518055, China.

$^{4}$Pazhou Lab, Guangzhou, 510330, China.

$^{5}$Harbin Institute of Technology, 92 West Dazhi Street, Nan Gang District, Harbin, 150001, China.

}
}

\maketitle
\vspace{-0.4in}
\begin{abstract}
The technology of dynamic map fusion among networked vehicles has been developed to enlarge sensing ranges and improve sensing accuracies for individual vehicles. This paper proposes a federated learning (FL) based dynamic map fusion framework to achieve high map quality despite unknown numbers of objects in fields of view (FoVs), various sensing and model uncertainties, and missing data labels for online learning. The novelty of this work is threefold: (1) developing a three-stage fusion scheme to predict the number of objects effectively and to fuse multiple local maps with fidelity scores; (2) developing an FL algorithm which fine-tunes feature models (i.e., representation learning networks for feature extraction) distributively by aggregating model parameters; (3) developing a knowledge distillation method to generate FL training labels when data labels are unavailable.
The proposed framework is implemented in the Car Learning to Act (CARLA) simulation platform. Extensive experimental results are provided to verify the superior performance and robustness of the developed map fusion and FL schemes.
\end{abstract}

\vspace{0.1in}

\IEEEpeerreviewmaketitle
\section{Introduction}

\begin{figure}[t]
\centering
\includegraphics[width=0.49\textwidth]{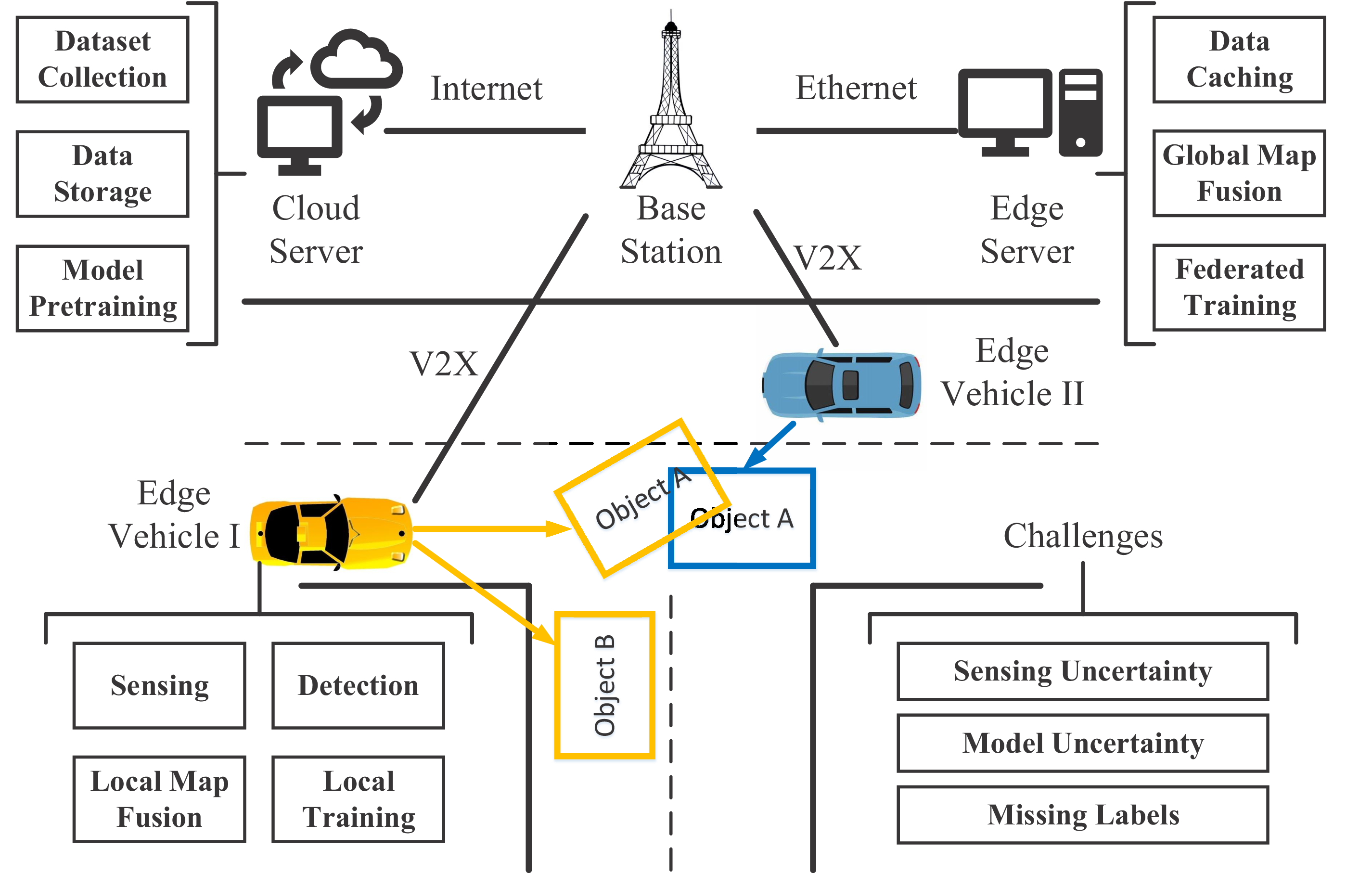}
\caption{An illustration of the intelligent networked vehicle system and three challenges for dynamic map fusion.}
\label{Fig.main}
\end{figure}

The intelligent networked vehicle system (INVS) is an emerging vehicle-edge-cloud system that accomplishes cooperative perception, map management, planning and maneuvering tasks via vehicle-to-everything (V2X) communication\cite{lu2014connected,eskandarian2019research,Shorinwa2020Distributed}. Among all the tasks, distributed map management aims to enlarge sensing ranges and improve sensing accuracies for individual vehicles and plays a central role in INVSs \cite{arnold2019cooperative,chen2019f,marvasti2020cooperative,wang2020v2vnet,xiao2018multimedia,chen2019cooper,miller2020cooperative,hurl2019trupercept,ambrosin2019object,yoon2019cooperative,yee2018collaborative,rawashdeh2018collaborative}. While static maps describe stationary objects (e.g., roads, buildings, and trees), dynamic maps
emphasize updating information of mobile objects (e.g., pedestrians, cars, and animals) in real time. Fig.~\ref{Fig.main} shows the architecture of an intelligent networked vehicle system. There are three main steps for dynamic map fusion: (1) local sensing and perception, (2) local map fusion and uploading, and (3) global map fusion and broadcasting.

\begin{table*}[!t]
\centering
\caption{A Comparison of Dynamic Map Fusion Schemes for Networked Intelligent Vehicles}
\vspace{0.1in}
{
\centering
\begin{tabular}{|c|c|c|c|c|c|c|c|c|}
\hline
\hline
\multirow{3}{*}{\makecell[c]{\textbf{Scheme}}} & \multirow{3}{*}{\centering \textbf{Literature}}                         & \multicolumn{3}{c|}{\textbf{Sensing and Fusion}}                                                                                                                                                            & \multicolumn{2}{c|}{\textbf{Perception Model}}                                                                                             & \textbf{Data Label}                                               & \multirow{3}{*}{\textbf{Limitation}}                                                                                      \\ \cline{3-8}
                                                                                     &                                                              & \textbf{\begin{tabular}[c]{@{}c@{}}Sensor \\ Modality\end{tabular}} & \textbf{\begin{tabular}[c]{@{}c@{}}Comm. \\ Overhead\end{tabular}} & \textbf{\begin{tabular}[c]{@{}c@{}}Weighted\\Fusion\end{tabular}} & \textbf{\begin{tabular}[c]{@{}c@{}}Deep\\  Networks\end{tabular}} & \textbf{\begin{tabular}[c]{@{}c@{}}Federated \\ Training\end{tabular}} & \textbf{\begin{tabular}[c]{@{}c@{}}Online  \\ Generation\end{tabular}} &                                                                                                                           \\ \hline
\textbf{\begin{tabular}[c]{@{}c@{}}Data \\ Level\end{tabular}}                     &\cite{chen2019cooper, arnold2019cooperative}                        & LiDAR                                                               & +++                                                                & \Checkmark                                                          & \Checkmark                                                                 & \XSolidBrush                                                                      & \XSolidBrush                                                                    & \begin{tabular}[c]{@{}c@{}}Demands for high communication
\\ workload and data synchronization\end{tabular}                      \\ \hline
\multirow{2}{*}{\textbf{\begin{tabular}[c]{@{}c@{}}Feature \\ Level\end{tabular}}} & \cite{chen2019f,marvasti2020cooperative,wang2020v2vnet}             & LiDAR                                                               & ++                                                                 & \Checkmark                                                          & \Checkmark                                                                 & \XSolidBrush                                                                      & \XSolidBrush                                                                    & \multirow{2}{*}{\begin{tabular}[c]{@{}c@{}}Demands for consistent feature\\ extraction and data synchronization
\end{tabular}} \\ \cline{2-8}
                                                                                     & \cite{xiao2018multimedia}                                           & Camera                                                              & ++                                                                 & \Checkmark                                                         & \Checkmark                                                                 & \Checkmark*                                                                     & \Checkmark*                                                                   &                                                                                                                           \\ \hline
\multirow{8}{*}{\textbf{\begin{tabular}[c]{@{}c@{}}Object \\ Level \end{tabular}}}  &
\begin{tabular}[c]{@{}c@{}}\cite{miller2020cooperative,ambrosin2019object}\\ \cite{yoon2019cooperative}
\end{tabular} & GPS                                                               & +                                                                  & \XSolidBrush                                                                & \XSolidBrush                                                                 & \XSolidBrush                                                                      & \XSolidBrush                                                                    & \begin{tabular}[c]{@{}c@{}}Impractical assumptions on \\ detection models
\end{tabular}                   \\ \cline{2-9}
                                                                                     & \cite{arnold2019cooperative,hurl2019trupercept}       & LiDAR                                                               & +                                                                  & \XSolidBrush                                                                & \Checkmark                                                                 & \XSolidBrush                                                                      & \XSolidBrush                                                                    & \begin{tabular}[c]{@{}c@{}}Requirement for high-quality
 \\ pretrained feature models\end{tabular}                                        \\ \cline{2-9}
                                                                                     & \cite{yee2018collaborative,rawashdeh2018collaborative}              & Camera                                                              & +                                                                  & \begin{tabular}[c]{@{}c@{}}\XSolidBrush  \end{tabular}       & \Checkmark                                                                 & \Checkmark*                                                                     & \Checkmark*                                                                   & \begin{tabular}[c]{@{}c@{}}Low accuracy of object \\ location and orientation
\end{tabular}                                                                                  \\ \cline{2-9}
                                                                                     & \textbf{Ours}                                                & LiDAR                                                               & +                                                                  & \begin{tabular}[c]{@{}c@{}}\Checkmark \end{tabular}          & \Checkmark                                                                 & \Checkmark                                                                      & \Checkmark                                                                    & \begin{tabular}[c]{@{}c@{}}Assumptions on high accuracy \\ of ego-vehicle localization\end{tabular}                                         \\ \hline
\hline
\end{tabular}
}
\vspace{0.1in}
\label{Table.related_work}

The symbol ``\checkmark'' means functionality available, ``\XSolidBrush'' means functionality not available, ``\checkmark*'' means functionality not available but supported. The number of ``+''s represents the level of communication overhead, more ``+'' s means higher overhead.
\end{table*}

Despite many efforts and successes in developing dynamic map fusion techniques \cite{arnold2019cooperative,chen2019f,marvasti2020cooperative,wang2020v2vnet,xiao2018multimedia,chen2019cooper,miller2020cooperative,hurl2019trupercept,yoon2019cooperative,ambrosin2019object,yee2018collaborative,rawashdeh2018collaborative},  a number of technical challenges still need to be properly handled, including
\begin{itemize}
    \item[1)] \textbf{Reduction of sensing uncertainties}. The sensing data at each individual vehicle may be missing, noisy, or mistaken due to sensor limitation and environmental complexity. A proper use of the redundant measurements from other vehicles can help reduce such uncertainties.
    \item[2)] \textbf{Reduction of model uncertainties}.
    The construction of dynamic maps at vehicles or edge servers relies on the quality of feature models of mobile objects. Those feature models, trained with labeled datasets at the cloud, may not be perfectly optimized due to many factors.
    \item[3)] \textbf{Missing labels for online learning}.
    While the datasets used at the cloud have been manually annotated, new data samples collected at vehicles are usually unlabeled. Those local sensory data samples should be automatically labeled for high-quality dynamic map updating.
\end{itemize}

Current dynamic map fusion techniques can be classified as data level \cite{chen2019cooper,arnold2019cooperative}, feature level\cite{chen2019f,marvasti2020cooperative,wang2020v2vnet,xiao2018multimedia}, and object level\cite{chen2019cooper,miller2020cooperative,hurl2019trupercept,yoon2019cooperative,ambrosin2019object,yee2018collaborative,rawashdeh2018collaborative}. Those data-level and feature-level methods likely incur high communication overheads and most object-level methods do not take uncertainties into account. On the other hand, various deep and federated learning schemes have been used to improve the quality of feature models\cite{xiao2018multimedia,yee2018collaborative,rawashdeh2018collaborative}, but few of them use point-cloud data. Besides, as most cloud-based model training schemes assume that datasets have been properly labeled\cite{chen2019cooper,arnold2019cooperative,chen2019f,marvasti2020cooperative,wang2020v2vnet,xiao2018multimedia,hurl2019trupercept,yee2018collaborative,rawashdeh2018collaborative}, knowledge distillation (KD) methods\cite{liu2020federated,liu2019lifelong,zhuang2020performance} need further investigation for label generation among networked vehicles.

In this paper, we propose an FL based dynamic map fusion framework, which enables scored-based object-level fusion and distributed online learning to achieve high map quality and low communication overhead. The main contributions of this work are summarized as follows.
\begin{itemize}
    \item[1)]Developing a three-stage map fusion, including the density based spatial clustering of applications with noise (DBSCAN), the score-based weighted average, and the intersection over union (IoU) based box pruning, to achieve global map fusion at the edge server.

    \item[2)]Developing a point-cloud based distributed FL algorithm, which fine-tunes feature models of mobile objects distributively by aggregating model parameters.

    \item[3)]Developing a KD method to transfer the knowledge from the edge server to individual vehicles. The labels of real-time sensory data are provided by road-side units or generated from the  three-stage fusion algorithm at the edge server.

    \item[4)]Implementation of the proposed FL-based dynamic map fusion framework in the CARLA simulation platform\cite{dosovitskiy2017carla} with extensive testing experiments in terms of developed evaluation benchmark metrics (the open-source codes are available at https://github.com/zijianzhang/CARLA\_INVS).
\end{itemize}

\begin{figure*}[!t]
 \centering
    \includegraphics[width=1\textwidth]{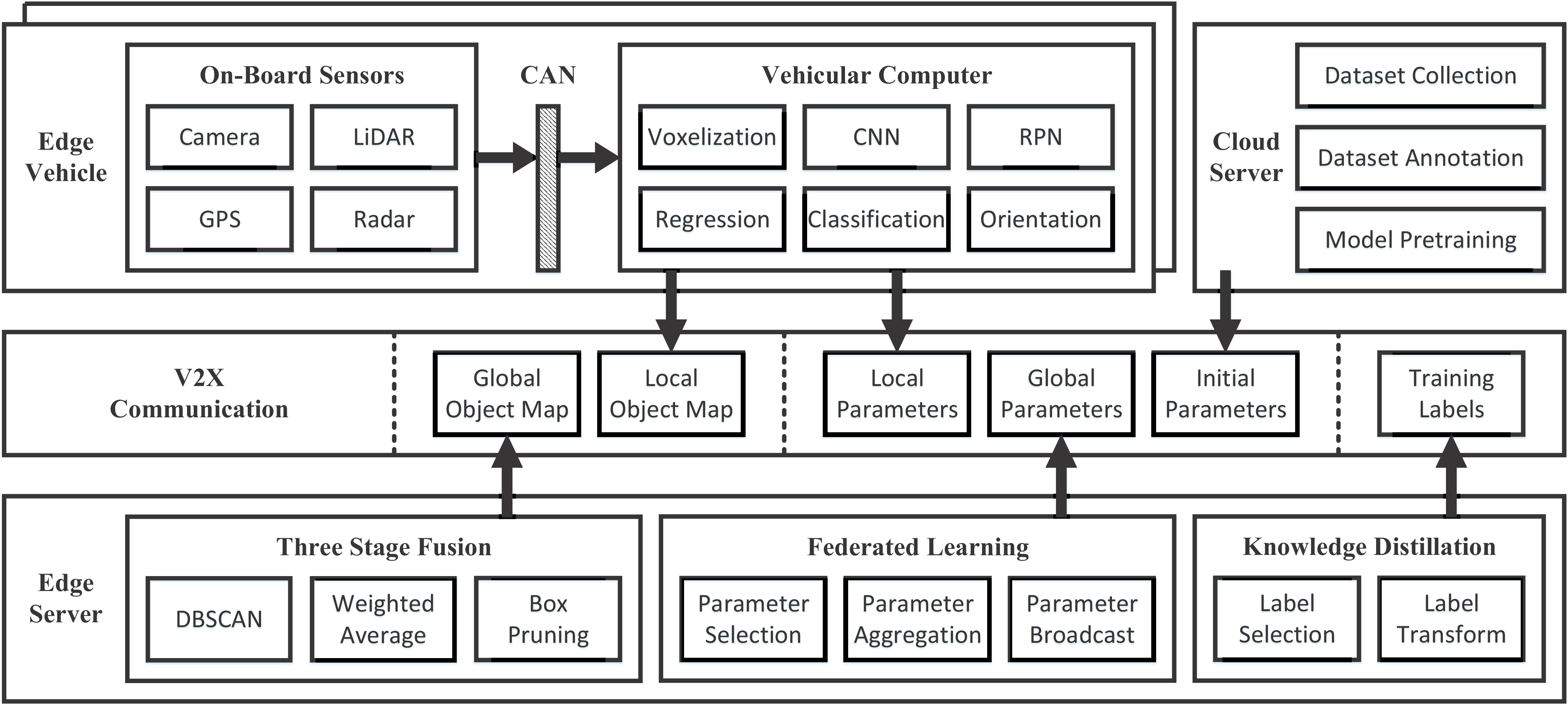}
  \caption{The proposed FL-based dynamic map fusion framework, which contains edge vehicles to generate local maps, a cloud server to perform feature model pre-training, an edge server to perform global map fusion, federated learning and knowledge distillation.}
  \label{Fig.framework}
\end{figure*}

\section{Related Work}

As summarized in Table.~\ref{Table.related_work}, current dynamic map fusion techniques can be categorized into data level \cite{chen2019cooper,arnold2019cooperative}, feature level\cite{chen2019f,marvasti2020cooperative,wang2020v2vnet,xiao2018multimedia}, and object level\cite{chen2019cooper,miller2020cooperative,hurl2019trupercept,yoon2019cooperative,ambrosin2019object,yee2018collaborative,rawashdeh2018collaborative}.
Data level methods collect raw data (such as point clouds) from different vehicles and achieve global dynamic map fusion by using deep networks\cite{chen2019cooper,arnold2019cooperative} at the cost of high communication overhead and extra data synchronization.
Feature level methods deploy identical feature extraction models (such as voxelization \cite{chen2019f,marvasti2020cooperative,wang2020v2vnet} and segmentation\cite{xiao2018multimedia}) at different vehicles, and achieve global dynamic map fusion via feature collection and feature processing based on deep networks.
Object level methods usually share the object lists among vehicles and achieve global dynamic map fusion through object association and multi-stage fusion procedures \cite{chen2019cooper,miller2020cooperative,hurl2019trupercept,yoon2019cooperative,ambrosin2019object,yee2018collaborative,rawashdeh2018collaborative}.

The object lists from different vehicles can be registered by using either ego-vehicle pose information\cite{yoon2019cooperative,ambrosin2019object,miller2020cooperative} or object geometric features\cite{arnold2019cooperative,hurl2019trupercept,yee2018collaborative,rawashdeh2018collaborative}. The former scheme needs constant communications among all ego-vehicles to estimate the transforms between any two poses and relies on impractical assumptions on detection models.
The later scheme relies more on the quality of object pose and size information predicted by each connected ego-vehicles and resolves conflicts via max-score fusion \cite{arnold2019cooperative} or mean fusion \cite{hurl2019trupercept}. Distance and IoU metrics have often been used to associate object lists, however leading to inaccurate estimation of the number or fused poses of objects\cite{ambrosin2019object,miller2020cooperative}.

Various deep learning techniques have been developed to improve the quality of object feature models\cite{chen2019cooper,arnold2019cooperative,chen2019f,marvasti2020cooperative,wang2020v2vnet,xiao2018multimedia,hurl2019trupercept,yee2018collaborative,rawashdeh2018collaborative}.
In particular, to learn from distributed data at different vehicles, federated learning (FL) schemes\cite{liu2020federated,liu2019lifelong,zhuang2020performance} have been developed for camera images and convolutional neural network based feature models.
This means that FL can be applied to image-based dynamic map fusion\cite{xiao2018multimedia,yee2018collaborative,rawashdeh2018collaborative}.
However, point cloud data can provide more accurate depth information, which have not been integrated with FL for dynamic maps.
On the other hand, deep network models trained at the cloud\cite{chen2019cooper,arnold2019cooperative,chen2019f,marvasti2020cooperative,wang2020v2vnet,xiao2018multimedia,hurl2019trupercept,yee2018collaborative,rawashdeh2018collaborative} assume that training datasets have been properly labeled, but real-world applications require label generation for new data samples during online learning procedures via knowledge transfer from ``teacher'' models to ``student'' models \cite{liu2020federated,liu2019lifelong,zhuang2020performance}.
Existing knowledge transfer either uses predictions of a single vehicle or average predictions of multiple vehicles as the ``teacher'' model, which may not fully exploit the useful information from all connected vehicles.

In this work, we propose an FL based dynamic map fusion framework, which enables scored-based object-level fusion and distributed online learning to achieve high map quality and low communication overhead. In the fusion process, we choose a density based clustering algorithm for object association and develop a dynamic weight adjustment algorithm to adaptively fuse object information. Furthermore, the object-level fusion results are chosen as the teacher model output for online federated learning. The complete framework of FL and KD based dynamic map fusion is developed based on multi-agent point-cloud datasets in CARLA.

\section{System Architecture and Algorithm Design}

We consider an environment which consists of $M$ dynamic objects and $K$ intelligent vehicles.
At the $t$-th time slot, the $m$-th object (with $1\leq m\leq M$) is represented by
\begin{align}
\mathbf{v}_m(t)=[c_m, x_m, y_m, z_m, l_m, w_m, h_m, \theta_m]^T, \label{vm}
\end{align}
where $c_m$ is the category, $(x_m, y_m, z_m)$ are the center coordinates, $(l_m, w_m, h_m)$ stand for the length, width, and height, and $\theta_m$ denotes the yaw rotation around the z-axis for the $m$-th object, respectively.
The global object map is the set of all objects $\mathcal{G}=\{\mathbf{v}_m(t)\}_{m=1}^M$, which is the output of our system.
To generate $\mathcal{G}$, this paper proposes an FL-based dynamic map fusion framework shown in Fig.~\ref{Fig.framework}.
There are three major components and one communication module as follows.

\textbf{Edge Vehicles}. At the $t$-th time slot and the $k$-th vehicle, the raw data (e.g., image, point cloud) is denoted as a vector $\mathbf{d}^{[k]}(t)\in\mathbb{R}^D$, where $D$ is the data dimension.
    The $k$-th vehicle fuses the raw data $\mathbf{d}^{[k]}(t)$ into an object list $\{\mathbf{u}^{[k]}_n(t),s^{[k]}_n(t)\}_{n=1}^{N_k}$ via a feature model $\Phi\left(\cdot|\mathbf{w}_k\right)$, where $N_k$ is the number of objects, $\mathbf{u}^{[k]}_n(t)$ is the $n$-th object, $s^{[k]}_n(t)$ is the object score representing the confidence of prediction, and $\mathbf{w}_k\in\mathbb{R}^W$ is the model parameter vector with model size $W$.
    The local fusion procedure is written as $\{\mathbf{u}^{[k]}_n(t),s^{[k]}_n(t)\}_{n=1}^{N_k}\leftarrow\Phi\left(\mathbf{d}^{[k]}(t)|\mathbf{w}_k\right)$.
    Object $\mathbf{u}^{[k]}_n(t)$ has the same format as $\mathbf{v}_m(t)$ in \eqref{vm}, but adopts local coordinate system of vehicle $k$.
    If labels $\{\mathbf{b}^{[k]}_n(t)\}$ for the $k$-th local object lists are available, then parameter $\mathbf{w}_k$ can be locally updated.
    The vehicle will upload its own location, $\{\mathbf{u}^{[k]}_n(t),s^{[k]}_n(t)\}_{n=1}^{N_k}$ and $\mathbf{w}_k$ to the edge server via V2X communication.
    This part is shown at the upper left of Fig.~\ref{Fig.framework}.

 \textbf{Cloud Server}. It performs dataset collection which gathers sensing data in various scenarios using dedicated vehicles, dataset annotation which labels the data manually, and model pretraining which outputs initial model parameters $\mathbf{w}^{[0]}\in\mathbb{R}^W$.
    The number of samples in each scenario is determined via learning curves \cite{lcpa,sun2020smart}.
    The cloud server will transmit $\mathbf{w}^{[0]}$ to edge server via the Internet and the edge server further broadcasts $\mathbf{w}^{[0]}$ to all vehicles.
    This part is shown at the upper right of Fig.~\ref{Fig.framework}.

 \textbf{Edge Server}. It performs three-stage fusion which generates global objects $\mathcal{G}$ by fusing the local objects $\{\mathbf{u}^{[k]}_n(t),s^{[k]}_n(t)\}_{n=1}^{N_k}$, federated learning which generates global model parameter vector $\mathbf{g}$ via aggregation of $\{\mathbf{w}_k\}_{k=1}^K$, and knowledge distillation which generates training labels $\{\mathbf{b}^{[k]}_n(t)\}$ either via teacher-student distillation \cite{kd} (e.g., road-side units provide labels) or ensemble distillation  \cite{ed} (e.g., global fusion provides labels).
    The edge server will broadcast $\mathcal{G}$, $\mathbf{g}$, and $\{\mathbf{b}^{[k]}_n(t)\}$ to all vehicles via V2X communication.
    This part is shown at the bottom of Fig.~\ref{Fig.framework}.

 \textbf{V2X Communication}. There are three data formats being exchanged via V2X: the object maps $\{\mathbf{u}^{[k]}_n(t),s^{[k]}_n(t)\}_{n=1}^{N_k}$ and $\mathcal{G}$ for the purpose of map fusion; the model parameters $\mathbf{g}$, $\mathbf{w}^{[0]}$, $\{\mathbf{w}_k\}_{k=1}^K$ for the purpose of federated learning; the labels $\{\mathbf{b}^{[k]}_n(t)\}$ for the purpose of knowledge distillation.

\begin{algorithm}[!t]
    \caption{Three-Stage Fusion}
        \begin{algorithmic}[1]
            \State \textbf{Input} local maps $\{\mathbf{u}^{[k]}_n(t),s^{[k]}_n(t)\}_{n=1}^{N_k}$
            \State \textbf{Output} global map $\mathcal{G}$, number of vehicles $M$, and association matrix $\{\mathbf{A}^{[k]}(t)\}$
            \State $\mathcal{X}=\{\{\mathbf{u}^{[1]}_n(t)\}_{n=1}^{N_1},\cdots,\{\mathbf{u}^{[K]}_n(t)\}_{n=1}^{N_K}\}$
            \State $\left(M,\{\mathbf{A}^{[k]}(t)\}_{k=1}^K\right)\leftarrow$DBSCAN$\left(F_{\rm{L\rightarrow G}} \left(\mathcal{X}\right)\right)$
            \State \textbf{For} each global object $m=1,\cdots,M$
            \State \ \ \ \textbf{For} each vehicle $k=1,\cdots,K$
            \State \ \ \ \ \ \ \textbf{For} each object $n=1,\cdots,N_k$
            \State \ \ \ \ \ \ \ \ \ \textbf{If} $a^{[k]}_{n,m}(t)=1$
            \State \ \ \ \ \ \ \ \ \ \ \ \ $\gamma^{[k]}_n(t)\leftarrow\frac{\left[1+\mathrm{exp}\left(s^{[k]}_n\right)\right]^{-1}}{\sum_{i=1}^K \sum_{j=1}^{N_i} a^{[i]}_{j,m}(t)\left[1+\mathrm{exp}\left(s^{[i]}_j(t)\right)\right]^{-1}}$
            \State \ \ \ \ \ \ \ \ \ \textbf{Else}
            \State \ \ \ \ \ \ \ \ \ \ \ \ $\gamma^{[k]}_n(t)\leftarrow0$
            \State \ \ \ \ \ \ \ \ \ \textbf{End}
            \State \ \ \ \ \ \ \textbf{End}
            \State \ \ \ \textbf{End}
            \State \ \ \  $\mathbf{v}_m(t)\leftarrow\sum_{k=1}^K\sum_{n=1}^{N_k}a^{[k]}_{n,m}(t)\gamma^{[k]}_n(t)\mathbf{u}^{[k]}_n(t)$
            \State \ \ \ $q_m(t)\leftarrow\sum_{k=1}^K\sum_{n=1}^{N_k}a^{[k]}_{n,m}(t)\gamma^{[k]}_n(t)s^{[k]}_n(t)$
            \State \textbf{End}
            \State \textbf{For} each object $m=1,\cdots,M$
            \State \ \ \ \textbf{For} each object $j=1,\cdots,M$
            \State \ \ \ \ \ \ \textbf{If} IoU$(\mathbf{v}_m(t),\mathbf{v}_j(t))>\delta$
            \State \ \ \ \ \ \ \ \ \ Remove $\mathbf{v}_m(t)$ if $q_m(t)<q_j(t)$
            \State \ \ \ \ \ \ \ \ \ Remove $\mathbf{v}_j(t)$ if $q_m(t)\geq q_j(t)$
            \State \ \ \ \ \ \ \textbf{End}
            \State \ \ \ \textbf{End}
            \State \textbf{End}
        \end{algorithmic}
\end{algorithm}

The above dynamic map fusion framework consists of three novel techniques: 1) three-stage fusion which tackles sensing uncertainties; 2) federated learning which tackles model uncertainties; 3) and knowledge distillation which tackles missing data labels. These techniques are summarized in the tables of Algorithm 1, Algorithm 2, and Algorithm 3.

\textbf{Algorithm 1: Three-Stage Fusion}. The first stage partitions the objects in local maps. Most association methods are sensitive to data with heterogeneous densities and hence unsuitable for dynamic map fusion, where the input data could be dense and complicated. We choose the DBSCAN method\cite{dbscan}, which can deal with unbalanced clusters and outliers pretty well. The output of this stage is the predicted number of objects $M$ and the association matrices $\{\mathbf{A}^{[k]}(t)\in\{0,1\}^{N_k\times M}\}$, where the element at the $n$-th row and $m$-th column is denoted as $\{a^{[k]}_{n,m}(t)\}$.  If the $n$-th object in the local map of vehicle $k$ is associated with the $m$-th object in the global map, then $a^{[k]}_{n,m}(t)=1$; otherwise $a^{[k]}_{n,m}(t)=0$. The second stage generates the objects in the global map. To account for the uncertainties at different vehicles, a score-based weighted average method is proposed, which solves the following weighted least squares problem:
\begin{align}\label{fusion}
\mathop{\mathrm{min}}_{\substack{\{\mathbf{v}_m(t)\}}}~&\sum_{k=1}^K\sum_{n=1}^{N_k}
\gamma^{[k]}_n(t)
a^{[k]}_{n,m}(t)
\Bigg\|\mathbf{v}_m(t)-F_{\rm{L\rightarrow G}}\left(\mathbf{u}^{[k]}_n(t)\right)\Bigg\|^2,
\end{align}
where the weights $\gamma^{[k]}_n(t)$ are determined by the sigmoid function of prediction scores $\{s_{n}^{k}(t)\}$, and $F_{\rm{L\rightarrow G}}$ is the function transforming local coordinates to global coordinates (vice versa for $F_{\rm{G\rightarrow L}}$).
The third stage eliminates overlapped boxes, i.e., $\mathbf{v}_m(t)$ and $\mathbf{v}_j(t)$ with $m\neq j$ may occupy the same space. There are two possible cases: 1) one of the detected objects does not exist; 2) both objects exist, but our predictions of the objects are inaccurate. To accommodate both cases, we first compute the IoUs of all overlapped box groups. If the IoU is larger than a threshold $\delta$, the object with the largest score is reserved and other objects are removed from the map. Otherwise, all the objects are reserved.

\textbf{Algorithm 2: Federated Learning}.
FL aims to find a common model parameter vector such that the total training loss function is minimized:
\begin{align}\label{FL}
\mathop{\mathrm{min}}_{\substack{\mathbf{w}_1=\cdots=\mathbf{w}_K}}
\quad&\sum_{k=1}^K\sum_{t=T_1}^{T_2}
\Xi\left(\mathbf{w}_k,\mathbf{d}^{[k]}(t),\mathbf{b}^{[k]}_n(t)\right),
\end{align}
where $(T_1,T_2)$ are the starting time and finishing time of training frames, $\Xi(\cdot) = \beta_1 L_{\rm{class}} + \beta_2\left(L_{\rm{angle}} + L_{\rm{box}}\right) + \beta_3 L_{\rm{dir}}$, $L_{\rm{class}}$ is the classification loss related to $\{c^{[k]}_n\}$, $L_{\rm{angle}}$ is the smooth $l_1$ function related to $\{\theta^{[k]}_n\}$, $L_{\rm{box}}$ is the box regression loss function related to $\{x^{[k]}_n, y^{[k]}_n, z^{[k]}_n, l^{[k]}_n, w^{[k]}_n, h^{[k]}_n\}$, $L_{\rm{dir}}$ the soft max function related to $\{\theta^{[k]}_n\}$ (to distinguish opposite directions), and $(\beta_1,\beta_2,\beta_3)$ are tuning parameters.
The training of FL model parameter (i.e., solving \eqref{FL}) is a distributed and iterative procedure, where each iteration involves two steps: 1) updating the local parameter vectors $(\mathbf{w}_1,\cdots,\mathbf{w}_K)$ using $\left(\mathbf{d}^{[1]}(t),\{\mathbf{b}^{[1]}_n(t)\},\cdots,\mathbf{d}^{[K]}(t),\{\mathbf{b}^{[K]}_n(t)\}\right)$ at the vehicles $(1,\cdots,K)$, respectively; and 2) computing the global parameter vector $\mathbf{g}$ by aggregating $(\mathbf{w}_1,\cdots,\mathbf{w}_K)$ at the edge server.
The entire procedure is stopped until the maximum number of iterations $I_{\mathrm{max}}$ is reached.

\begin{algorithm}[!t]
    \caption{Federated Learning}
        \begin{algorithmic}[1]
            \State \textbf{Input} raw data $\mathbf{d}^{[k]}(t)$ and labels $\{\mathbf{b}^{[k]}_n(t)\}$ of vehicle $k$ from time $t=T_1$ to $t=T_2$
            \State \textbf{Output} global model parameter vector $\mathbf{g}$
            \State Initialize $\mathbf{g}^{[1]}=\mathbf{w}_1^{[1]}=\cdots=\mathbf{w}_K^{[1]}=\mathbf{w}^{[0]}$
            \State \textbf{For} each round $i=1,\cdots,I_{\mathrm{max}}$
            \State \ \ \ \textbf{For} each vehicle $k=1,\cdots,K$ \textbf{in parallel} do
            \State \ \ \ \ \ \ Split $\{\mathbf{d}^{[k]}(t), \mathbf{b}^{[k]}_n(t)\}_{t=T_1}^{T_2}$ into batches $\mathcal{B}=\{\mathcal{A}_1,\mathcal{A}_2,\cdots\}$
            \State \ \ \ \ \ \ \textbf{For} each local epoch $\tau=1,\cdots,E$
            \State \ \ \ \ \ \ \ \ \ \textbf{For} each batch $\mathcal{A}_j\in\mathcal{B}$
            \State \ \ \ \ \ \ \ \ \ \ \ \ $\mathbf{w}^{[i]}_k\leftarrow\mathbf{w}^{[i]}_k-\varepsilon\mathop{\sum}_{(\mathbf{d}^{[k]}(t),\mathbf{b}^{[k]}_n(t))\in\mathcal{A}_j}\nabla_{\mathbf{w}_k}\Xi\left(\mathbf{w}^{[i]}_k, \mathbf{d}^{[k]}(t),\mathbf{b}^{[k]}_n(t)\right)$
            \State \ \ \ \ \ \ \ \ \ \textbf{End}
            \State \ \ \ \ \ \ \textbf{End}
            \State \ \ \ \ \ \ Return $\mathbf{w}^{[i]}_k$ to the edge server
            \State \ \ \ \textbf{End}
            \State \ \ \ Server executes $\mathbf{g}^{[i+1]}=\mathbf{w}_1^{[i+1]}=\cdots=\mathbf{w}_K^{[i+1]}=\frac{1}{K}\sum_{k=1}^K\mathbf{w}_k^{[i]}$
            \State \textbf{End}
        \end{algorithmic}
\end{algorithm}
\begin{algorithm}[!t]
    \caption{Knowledge Distillation}
        \begin{algorithmic}[1]
            \State \textbf{Input} local maps $\{\mathbf{u}^{[k]}_n(t),s^{[k]}_n(t)\}_{n=1}^{N_k}$; global fusion map $\{\mathbf{v}_m\}$; association matrices $\{\mathbf{A}^{[k]}\}$; set of teachers $\mathcal{T}$ and teacher model $\{f_1,\cdots f_{|\mathcal{T}|}\}$; set of students $\mathcal{S}$
            \State \textbf{Output} labels $\{\mathbf{b}^{[k]}_n(t)\}$ for $k\in\mathcal{S}$
            \State Initialize $\mathbf{b}^{[k]}_n(t)=[~]$
            \State \textbf{For} each vehicle $k\in\mathcal{S}$
            \State \ \ \ \textbf{For} each object $n=1,\cdots,N_k$
            \State \ \ \ \ \ \ \textbf{For} each teacher $i\in\mathcal{T}$
            \State \ \ \ \ \ \ \ \ \ \textbf{If} $i$ has label of object $n$
            \State \ \ \ \ \ \ \ \ \ \ \ \ $\mathbf{b}^{[k]}_n(t)\leftarrow f_i(\mathbf{u}^{[k]}_n(t))$, where $f_i$ is the teacher model
            \State \ \ \ \ \ \ \ \ \ \ \ \ Continue
            \State \ \ \ \ \ \ \ \ \ \textbf{End}
            \State \ \ \ \ \ \ \textbf{End}
            \State \ \ \ \ \ \ \textbf{If} $\mathbf{b}^{[k]}_n(t)=[~]$
            \State \ \ \ \ \ \ \ \ \ $\mathbf{b}^{[k]}_n(t)\leftarrow F_{\rm{G\rightarrow L}}\left(\sum_{m=1}^Ma_{n,m}^{[k]}\mathbf{v}_m\right)$
            \State \ \ \ \ \ \ \textbf{End}
            \State \ \ \ \textbf{End}
            \State \textbf{End}
        \end{algorithmic}
\end{algorithm}

\begin{figure*}[!t]
\centering
\subfigure[]{
\label{Fig.fusion.1}
\includegraphics[width=0.23\textwidth]{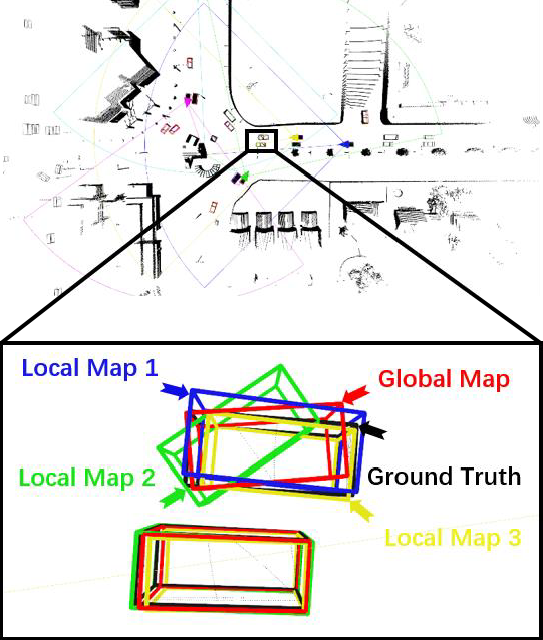}}
\subfigure[]{
\label{Fig.fusion.2}
\includegraphics[width=0.23\textwidth]{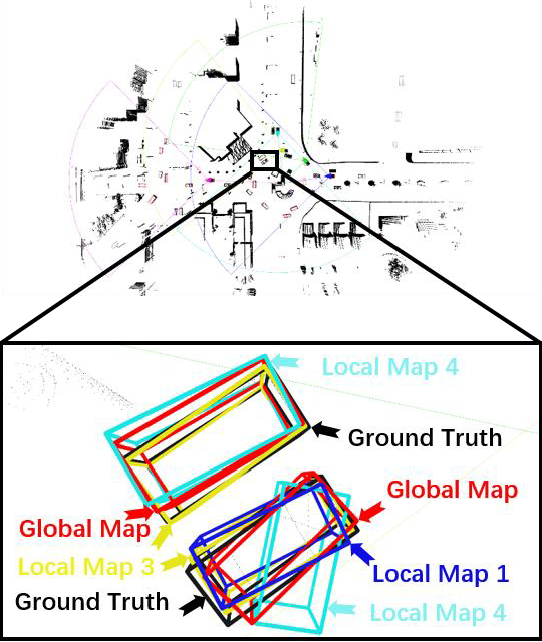}}
\subfigure[]{
\label{Fig.fusion.3}
\includegraphics[width=0.23\textwidth]{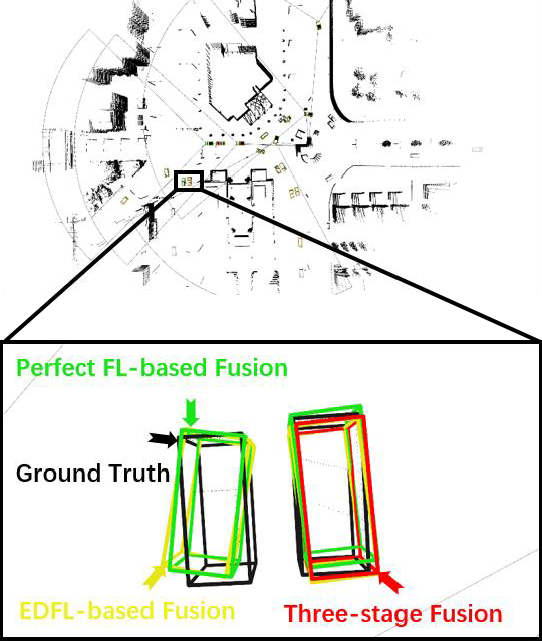}}
\subfigure[]{
\label{Fig.fusion.4}
\includegraphics[width=0.23\textwidth]{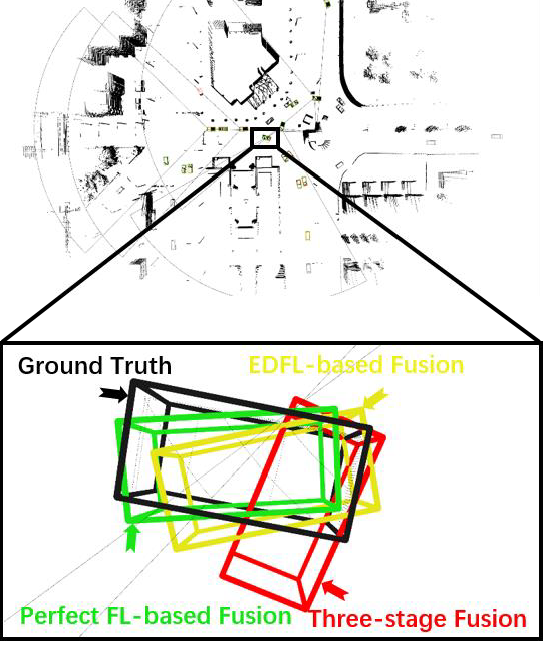}}
\caption{
a) and b) An illustration of dynamic maps with and without fusion. Black box: ground truth. Red box: fusion results. Other colored boxes: local detection results. c) and d) An illustration of the three stage fusion, perfect FL-based fusion, and EDFL-based fusion results. Black box: ground truth. Red box: prediction with three stage fusion. Green box: prediction with perfect FL-based fusion. Yellow box: prediction with EDFL-based fusion.}
\label{Fig.fusion}
\end{figure*}

\textbf{Algorithm 3: Knowledge Distillation}.
KD aims to generate training labels for vehicles in the student set $\mathcal{S}$.
The student set is determined as follows: if the vehicle finds that its own local map and the global map differ too much, it will be selected into $\mathcal{S}$.
We consider two types of KD:
teacher-student distillation and ensemble distillation.
For the first type, it is assumed that there exists a teacher model that can produce ground truth labels.
For example, some objects are intelligent vehicles that can upload their locations and orientations;
or the road side units provide labels in their FoVs.
For the second type, the global map, which fuses the information from all local maps, is used to produce training labels.
Notice that leveraging the consensus between vehicles (fusion) could potentially add more information, but this will inevitably lead to higher bias (bias-variance trade-off).
However, as shown in the experimental results, this type of KD also improves the system performance.
The integration of FL and this type of KD is termed ensemble distillation federated learning (EDFL).

\section{Experimental Results}

\begin{figure*}[!t]
\vspace{-0.3cm}
\centering
\subfigure[Vehicle 1]{
\label{Fig.fed.1}
\includegraphics[width=0.185\textwidth]{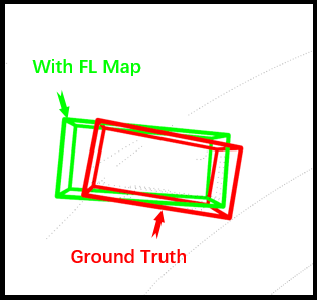}}
\subfigure[Vehicle 2]{
\label{Fig.fed.2}
\includegraphics[width=0.185\textwidth]{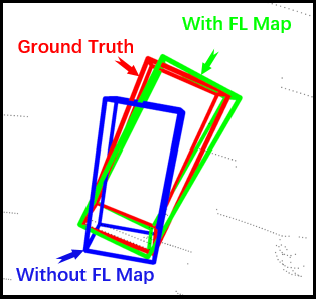}}
\subfigure[Vehicle 3]{
\label{Fig.fed.3}
\includegraphics[width=0.185\textwidth]{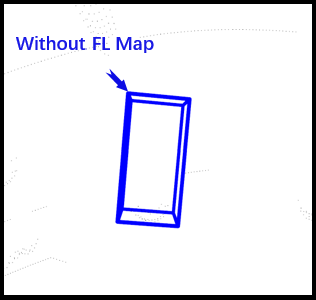}}
\subfigure[Vehicle 4]{
\label{Fig.fed.4}
\includegraphics[width=0.185\textwidth]{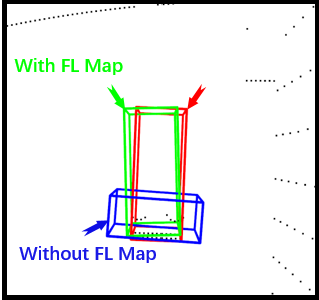}}
\subfigure[Vehicle 5]{
\label{Fig.fed.5}
\includegraphics[width=0.185\textwidth]{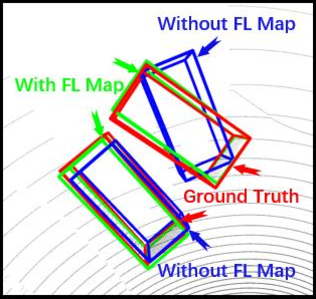}}
\caption{An illustration of dynamic maps with or without FL. Red box: ground truth. Blue box: predictions without FL. Green box: predictions with FL.}
\label{Fig.fed}
\end{figure*}

\begin{figure*}[!t]
\centering
\subfigure[Object Maps Average Precision at IoU=0.7]{
\label{Fig.fusion_table}
\includegraphics[height=0.27\textwidth]{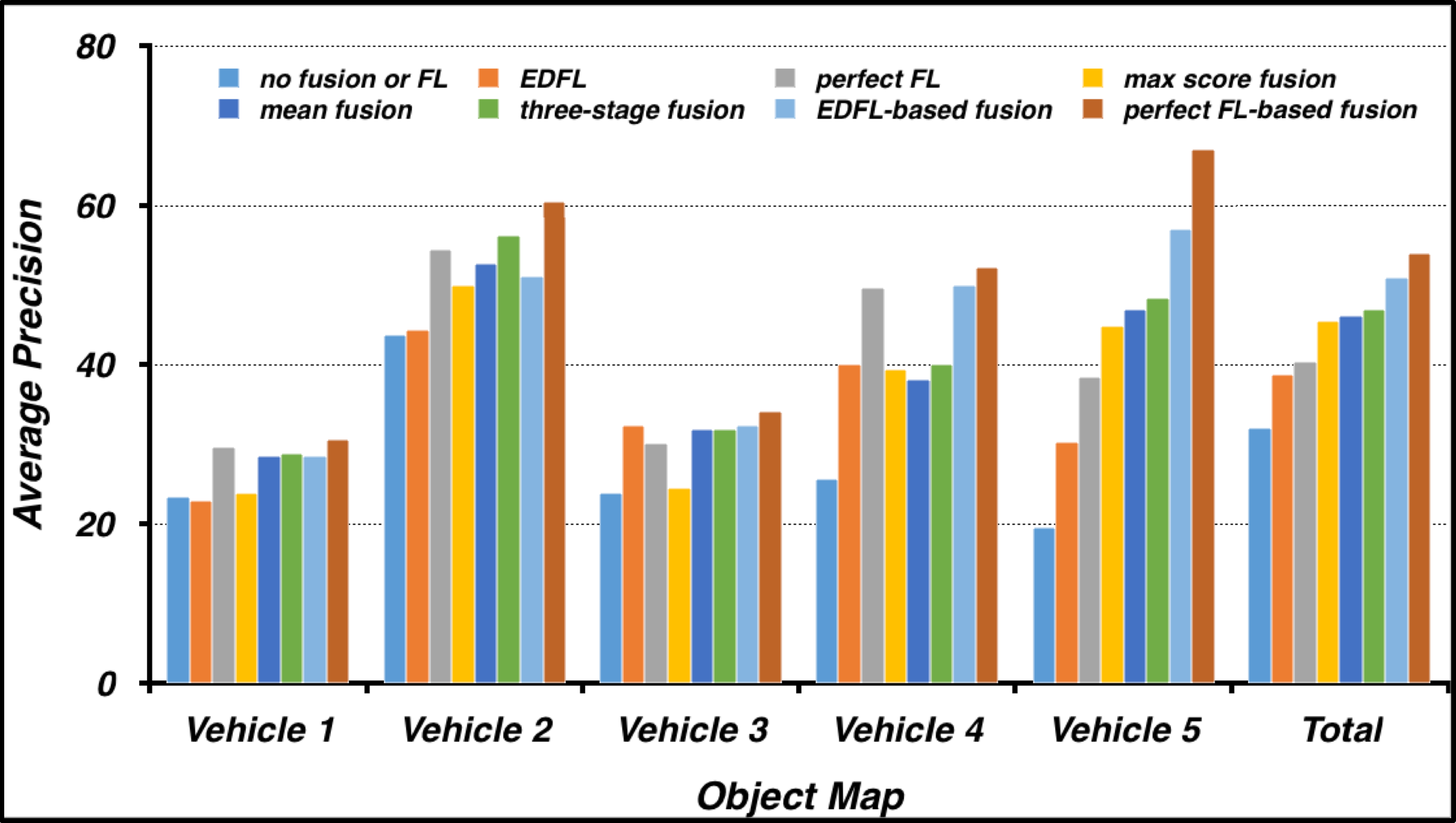}}
\subfigure[Methods Performance at IoU=0.7]{
\label{Fig.federated_table}
\includegraphics[height=0.27\textwidth]{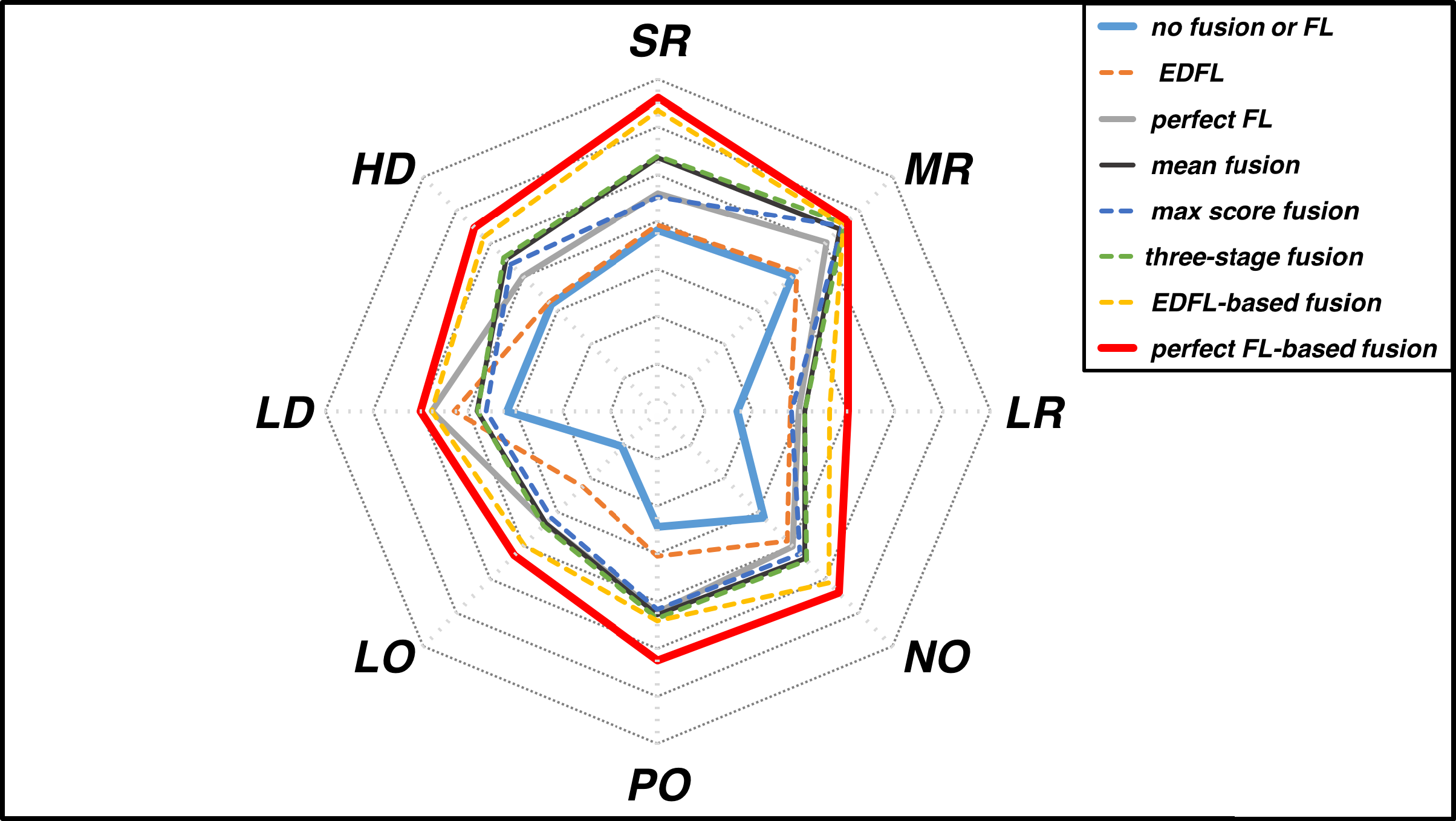}}
\label{Fig.performance}
\caption{a) A comparison of different map fusion schemes in INVS at $0.7$ IoU for individual vehicles and all vehicles. b) A comparison of map fusion schemes in INVS between the proposed method and other methods in terms of proposed benchmark metrics (SR, MR, LR, NO, PO, LO, LD, HD) at IoU$=0.7$ , where the perfect FL-based fusion yields the upper bound of map fusion performance when there are no missing data labels.}
\end{figure*}

We employ the CARLA simulation platform to generate training and testing scenarios and multi-agent point cloud datasets for INVS. Each intelligent vehicle (Tesla Model 3) is equipped with a 64-line LiDAR at 20 Hz and a GPS device. The default LiDAR range is set to $100$ m, and its FoV is $90$ degree on front. Since the sensing data generated by CARLA is not compatible to the SECOND network\cite{yan2018second,shi2020points}, we develop a data transformation module such that the generated dataset satisfies the KITTI standard \cite{geiger2013vision}.

In the pretraining stage, three intelligent vehicles with different traveling routes are employed to collect sensing data in the `Town03' map of CARLA. The pre-training dataset includes $60000$ frames of point clouds, and 900 frames are chosen for training object feature models. The Adam optimizer is adopted with a learning rate ranging from $10^{-4}$ to $10^{-6}$.
The number of epochs is set to $50$. After the pretraining stage, the parameters of all intelligent vehicles are set to $\mathbf{w}_1=\cdots\mathbf{w}_K=\mathbf{w}^{[0]}$.

Then we simulate a crossroad traffic scenario in case of $K = 5$ and $M = 37$, that is, $5$ intelligent vehicles and $32$ ordinary vehicles.
The entire scenario lasts for $50.5$ seconds and contains $1010$ frames. The first 510 frames are used for FL and KD (i.e., $T_1=0$ and $T_2=25.5$ in problem \eqref{FL}).
The sampling rate is chosen as $3:1$ and hence $170$ frames are used for training. The $500$ frames from $T_2 = 25.5$ to $T_3 = 50.5$ are used for inference, fusion, and testing. For FL, the number of local updates is set to $E=2$ and the total number of FL iterations $I_{\rm{max}}=5$.
We adopt the same training optimizer and learning rates as the cloud pretraining procedure.
The average precision at IoU$=0.7$ is used for performance evaluation.
We consider both perfect FL and EDFL algorithms.
For perfect FL algorithm, 170 training samples from $T_1$ to $T_2$ are perfectly labeled by a teacher model.
For EDFL algorithm, 170 training samples from $T_1$ to $T_2$ are not labeled.
The edge aggregates $170$ frames of the global map via the three-stage fusion.
All the vehicles use the global map to obtain their training labels, and the $170$ samples with the imperfect labels are used for training.

We develop a set of performance evaluation benchmark metrics for sensing ranges, occlusion rates, and traffic densities, including short range (SR), middle range (MR), long range (LR), no occlusion (NO), partial occlusion (PO), large amounts of occlusion (LO), low density(LD) and high density (HD).
Objects are classified into SR, MR, and LR according to the object-to-vehicle distances.
Objects are classified into NO, PO, and LO according to the number of point clouds reflected from objects within a certain sensing range.
A frame is labeled as HD if at least one object in that frame is detected by $\geq3$ intelligent vehicles.

Fig. \ref{Fig.fusion.1} and Fig. \ref{Fig.fusion.2} illustrate the performance of three stage fusion.
Fig. \ref{Fig.fusion.1} shows that despite the green vehicle generating false orientations, the global map (i.e., the red box) can correct the orientation error for the green vehicle.
Fig. \ref{Fig.fusion.2} shows that despite the blue vehicle missing one object, the global map (i.e., the red box) can recover that object for the blue vehicle.
Fig. \ref{Fig.fed} illustrates the performance of FL without fusion.
It can be seen that FL recovers the missing objects, and correct false orientation and false positive detection.
Fig. \ref{Fig.fusion.3} and Fig. \ref{Fig.fusion.4} illustrate the performance of FL-based fusion.
It can be seen that with imperfect labels, the EDFL-based fusion outperforms the three-stage fusion and achieves performance close to the perfect FL-based fusion.
Fig. \ref{Fig.fusion_table} compares the scheme without fusion or FL, the EDFL, the perfect FL, the mean fusion \cite{arnold2019cooperative}, the max score fusion \cite{hurl2019trupercept}, the three-stage fusion, the perfect FL-based fusion, and the EDFL-based fusion for individual vehicles and all vehicles.
Compared with max score fusion and mean fusion, the proposed three-stage fusion algorithm achieves the highest precision scores (when IoU$=0.7$).
Compared with the scheme without fusion or FL, the proposed perfect FL and EDFL algorithms can help increase the average precision scores.
Compared with all the other methods. the proposed perfect FL-based fusion and the EDFL-based fusion achieve much better performance.
Fig.~\ref{Fig.federated_table} shows that the perfect FL-based fusion scheme and the scheme with no fusion or FL achieve the most outer and inner lines, respectively.
The proposed three-stage fusion algorithm achieves the best performance for all the benchmark metrics among all fusion methods.
The performance of the proposed EDFL-based fusion algorithm is close to that of perfect FL-based fusion, especially for LR and LO metrics.

\section{Conclusion}

This paper presented an FL-based dynamic map fusion framework, which achieves high quality map fusion and low communication overhead.
The FL and KD methodologies were developed to achieve distributive and online feature model updating, as well as to consider common uncertainties that occur in real-world applications.
Experimental results based on multi-agent simulations in CARLA demonstrated that even without data labels, the proposed FL-based dynamic map fusion algorithm outperforms other existing methods in terms of the proposed benchmark metrics in INVSs.
Such a framework can be further extended to distributed static semantic map fusion.


\begin{thebibliography}{60}

\bibitem{lu2014connected} N.~Lu, N.~Cheng, N.~Zhang, X.~Shen, and J.~W.~Mark, ``Connected vehicles: Solutions and challenges,'' \emph{IEEE Internet of Things Journal}, vol.~1, no.~4, pp.~289--299, Aug.~2014.

\bibitem{eskandarian2019research} A.~Eskandarian, C.~Wu, and C.~Sun, ``Research Advances and Challenges of Autonomous and Connected Ground Vehicles,'' \emph{IEEE Transactions on Intelligent Transportation Systems}, vol.~22, no.~2, pp.~683--711, Feb.~2021.

\bibitem{Shorinwa2020Distributed} O.~Shorinwa, J.~Yu, T.~Halsted, A.~Koufos, and M.~Schwager, ``Distributed multi-target tracking for autonomous vehicle fleets,'' in \emph{Proceedings of 2020 IEEE International Conference on Robotics and Automation (ICRA)},   Paris, France, May--Aug.~2020, pp.~3495--3501.

\bibitem{arnold2019cooperative} E.~Arnold, M.~Dianati, R.~de~Temple, and S.~Fallah, ``Cooperative perception for 3D object detection in driving scenarios using infrastructure sensors,'' \emph{IEEE Transactions on Intelligent Transportation Systems}, early access, Oct.~2020. DOI: 10.1109/TITS.2020.3028424.

\bibitem{chen2019f} Q.~Chen, X.~Ma, S.~Tang, J.~Guo, Q.~Yang, S.~Fu, ``F-cooper: feature based cooperative perception for autonomous vehicle edge computing system using 3D point clouds,'' in \emph{Proceedings of 2019 ACM/IEEE Symposium on Edge Computing}, Arlington, Virginia, Nov.~2019, pp.~88--100.

\bibitem{marvasti2020cooperative} E.~E.~Marvasti, A.~Raftari, A.~E.~Marvasti, Y.~P.~Fallah,  R.~Guo, and H.~Lu, ``Cooperative LiDAR object detection via feature sharing in deep networks,'' 2020. [Online]. Available: arxiv preprint arXiv: 2002.08440.

\bibitem{wang2020v2vnet} T.-H.~Wang, S.~Manivasagam, M.~Liang, B.~Yang, W.~Zeng, and R.~Urtasun, ``V2VNet: Vehicle-to-vehicle communication for joint perception and prediction,'' in \emph{Proceedings of 2020 European Conference on Computer Vision (ECCV)}, Glasgow, Scotland, Aug.~2020.

\bibitem{xiao2018multimedia} Z.~Xiao, Z.~Mo, K.~Jiang, and D.~Yang, ``Multimedia fusion at semantic level in vehicle cooperactive perception,'' in \emph{Proceedings of 2018 IEEE International Conference on Multimedia \& Expo Workshops (ICMEW)}, San Diego, USA, Jul.~2018, pp.~1--6.

\bibitem{chen2019cooper} Q.~Chen, S.~Tang, Q.~Yang, and S.~Fu, ``Cooper: Cooperative perception for connected autonomous vehicles based on 3d point clouds,'' in \emph{Proceedings of 2019 IEEE International Conference on Distributed Computing Systems (ICDCS)}, Dallas, TX, 2019, pp.~514--524.

\bibitem{miller2020cooperative} A.~Miller, K.~Rim, P.~Chopra, P.~Kelkar, and M.~Likhachev, ``Cooperative perception and localization for cooperative driving,'' in \emph{Proceedings of 2020 IEEE International Conference on Robotics and Automation (ICRA)}, Paris, France, May--Aug.~2020, pp.~1256--1262.

\bibitem{hurl2019trupercept} B.~Hurl, R.~Cohen, K.~Czarnecki, and S.~Waslander, ``TruPercept: Trust modelling for autonomous vehicle cooperative perception from synthetic data,'' 2019. [Online]. Available: arXiv preprint arXiv:1909.07867.

\bibitem{ambrosin2019object} M.~Ambrosin, I. J.~Alvarez, C.~Buerkle, L.~L.~Yang, F.~Oboril, M. R.~Sastry, and K.~Sivanesan, ``Object-level perception sharing among connected vehicles,'' in \emph{Proceedings of 2019 IEEE Intelligent Transportation Systems Conference (ITSC)}, Auckland, New Zealand, Oct.~2019, pp.~1566--1573.

\bibitem{yoon2019cooperative} D.~D.~Yoon, G. M. N.~Ali, and B.~Ayalew, ``Cooperative perception in connected vehicle traffic under field-of-view and participation variations,'' in \emph{Proceedings of 2019 IEEE Connected and Automated Vehicles Symposium (CAVS)}, Honolulu, HI, Sep.~2019, pp.~1--6.

\bibitem{yee2018collaborative} R.~Yee, E.~Chan, B.~Cheng, and G.~Bansal, ``Collaborative perception for automated vehicles leveraging vehicle-to-vehicle communications,'' in \emph{Proceedings of 2018 IEEE Intelligent Vehicles Symposium (IV)}, Changshu, China, Jun.~2018, pp.~1099--1106.

\bibitem{rawashdeh2018collaborative} Z.~Y.~Rawashdeh and Z.~Wang, ``Collaborative automated driving: A machine learning-based method to enhance the accuracy of shared information,'' in \emph{Proceedings of 2018 International Conference on Intelligent Transportation Systems (ITSC)}, Maui, HI, Nov.~2018, pp.~3961--3966.

\bibitem{liu2020federated} B.~Liu, L.~Wang, M.~Liu, and C.-Z.~Xu, ``Federated imitation learning: A novel framework for cloud robotic systems with heterogeneous sensor data,'' \emph{IEEE Robotics and Automation Letters}, vol.~5, no.~2, pp.~3509--3516, Apr.~2020.

\bibitem{liu2019lifelong} B.~Liu, L.~Wang, and M.~Liu, ``Lifelong federated reinforcement learning: A learning architecture for navigation in cloud robotic systems,'' \emph{IEEE Robotics and Automation Letters}, vol.~4, no.~4, pp.~4555--4562, Oct.~2019.

\bibitem{zhuang2020performance} W.~Zhuang, Y.~Wen, X.~Zhang, X.~Gan, D.~Yin, D.~Zhou, S.~Zhang, and S.~Yi, ``Performance optimization for federated person re-identification via benchmark analysis,'' 2020. [Online]. Available: arXiv preprint arXiv:2008.11560.

\bibitem{dosovitskiy2017carla} A.~Dosovitskiy, G.~Ros, F.~Codevilla, A.~Lopez, and V.~Koltun, ``CARLA: An open urban driving simulator,'' in \emph{Proceedings of the 1st Annual Conference on Robot Learning (CoRL)}, Mountain View, CA, Oct.~2017, pp.~1--16.

\bibitem{lcpa} S.~Wang, R.~Wang, Q.~Hao, Y.-C.~Wu, and H.~V.~Poor, ``Learning centric power allocation for edge intelligence,'' in \emph{Proceedings of 2020 IEEE International Conference on Communications (ICC)}, Dublin, Ireland, Jun. 2020.

\bibitem{sun2020smart} Y.~Sun, D.~Li, X.~Wu, and Q.~Hao, ``Visual Perception Based Situation Analysis of Traffic Scenes for Autonomous Driving Applications,'' in \emph{Proceedings of 2020 IEEE International Conference on Intelligent Transportation Systems (ITSC)}, Rhodes, Greece, Jul.~2020, pp.~1--7.

\bibitem{kd} Q.~Guo, X.~Wang, Y.~Wu, Z.~Yu, D.~Liang, X.~Hu, and P.~Luo, ``Online Knowledge Distillation via Collaborative Learning,'' in \emph{Proceedings of 2020 IEEE/CVF Conference on Computer Vision and Pattern Recognition (CVPR)}, Seattle, WA, Jun.~2020, pp.~11017--11029.

\bibitem{ed} T.~Lin, L.~Kong, S.~U.~Stich, and M.~Jaggi, ``Ensemble Distillation for Robust Model Fusion in Federated Learning,'' in \emph{Proceedings of 2020 Conference on Neural Information Processing Systems (NeurIPS)}, Vancouver, Canada, Dec.~2020, pp.~1--25.

\bibitem{dbscan} M.~Ester, H.~P.~Kriegel, J.~Sander, and X.~Xu, ``A density-based algorithm for discovering clusters in large spatial databases with noise,'' in \emph{Proceedings of the 2nd International Conference on Knowledge Discovery and Data Mining (KDD)}, Portland, Oregon, 1996, pp.~226--231.

\bibitem{yan2018second} Y.~Yan, Y.~Mao, and B.~Li, ``SECOND: Sparsely embedded convolutional detection,'' \emph{Sensors}, vol.~18, no.~10, pp.~3337, 2018.

\bibitem{shi2020points} S.~Shi, Z.~Wang, J.~Shi, X.~Wang, and H.~Li, ``From points to parts: 3D object detection from point cloud with part-aware and part-aggregation network,'' \emph{IEEE Trans. Pattern Anal. Mach. Intell.}, 2020. DOI: 10.1109/TPAMI.2020.2977026.

\bibitem{geiger2013vision}  A. Geiger, P. Lenz, C. Stiller, and R. Urtasun, ``Vision meets robotics: The kitti dataset,'' \emph{International Journal of Robotics Research}, vol. 32, no.~11, pp. 1231--1237, Aug. 2013.


\end{thebibliography}
\end{document}